\documentclass[11pt]{article}
\usepackage[final]{acl}

\usepackage{times}
\usepackage{latexsym}
\usepackage{amsmath, amsfonts, amssymb}
\usepackage{booktabs}
\usepackage{multirow}
\usepackage{xcolor, soul}
\usepackage{enumitem}
\usepackage{graphicx}
\usepackage{subcaption}
\usepackage{siunitx}
\usepackage[table]{xcolor}
\usepackage{caption}
\usepackage[T1]{fontenc}
\usepackage[utf8]{inputenc}
\usepackage{microtype}
\usepackage{inconsolata}

\usepackage{algorithm}
\usepackage{algpseudocode}

\usepackage{booktabs}
\usepackage{siunitx}
\usepackage[table]{xcolor}
\author{
Mohsen Nayebi Kerdabadi, Arya Hadizadeh Moghaddam, Dongjie Wang, Zijun Yao\thanks{Corresponding author.}\\
University of Kansas, USA \\
\texttt{\{mohsen.nayebi, a.hadizadehm, wangdongjie, zyao\}@ku.edu}
}

\title{REFINE: LLM Refinement over Budgeted Text-Attributed Graphs for Personalized Medical Concept Representation}

\begin{document}
\maketitle

\newcommand{\model}{\textsc{REFINE}}

\begin{abstract}
Learning rich medical concept representations is essential for EHR prediction. Text-attributed knowledge graphs (TKGs) provide a natural foundation by organizing heterogeneous medical relations together with textual semantics. However, most existing encoders process concepts uniformly across patients, despite the fact that a code's meaning and predictive value depend on patient-specific clinical context and trajectory. Learning patient-personalized concept representations from TKGs introduces two key challenges: (1) deciding how much KG context to incorporate for each observed code, and (2) aligning semantic information with the patient-specific relational structure. We propose \model{}, a KG-aware budgeted LLM graph refinement framework for patient-personalized medical concept encoding. Starting from a global TKG, \model{} constructs patient-specific temporal graphs. A sequential reinforcement learning policy selects a personalized KG expansion budget for each observed code. The resulting patient graph is processed by a heterogeneous GNN to capture relation-aware structural dependencies, while a frozen LLM uses graph-aware soft prompts to semantically refine concept representations. Experiments on MIMIC-III and MIMIC-IV show that \model{} consistently improves diverse EHR backbones, outperforms strong baselines, and demonstrates robust gains across component ablation, KG selection, and data insufficiency.
\end{abstract}

\section{Introduction}

Electronic health records (EHRs) provide a longitudinal view of patient health through high-dimensional sequences of diagnosis, medication, and procedure codes, enabling a wide range of clinical prediction tasks~\cite{nayebi2023contrastive,jiang2023graphcare}. A central challenge in EHR mining is to learn medical concept representations that capture both rich clinical semantics and heterogeneous relational dependencies among codes, such as treatment patterns and comorbidities. Text-attributed knowledge graphs (TKGs) provide a natural way to organize these semantic and relational signals by linking medical concepts through typed relations and textual attributes. However, most existing concept encoders learn codes in a static manner across patients, overlooking that the same concept can have different clinical meaning and predictive value depending on patient history, visit context, and treatment trajectory. The same diagnosis may reflect a stable chronic condition in one patient but signal acute deterioration in another; likewise, a medication may be routine in one trajectory but highly informative when paired with specific comorbidities or procedures. Thus, beyond learning general concept semantics from TKGs, medical concept representations should be personalized to each patient’s local clinical context. Personalizing concept representations, however, is non-trivial and raises two challenges.

The first challenge concerns the relational aspect of the TKG: how much KG context should be attached to each observed code in a patient's history. Existing KG-based EHR models~\cite{jiang2023graphcare, kerdabadi2026text} often apply fixed expansion rules, retrieving the same hop depth or neighbor budget for every code across patients. This uniform strategy ignores that the value of relational context varies across patients and concepts. A code may be self-explanatory in one trajectory but require richer multi-hop evidence in another to reveal comorbidities, treatment dependencies, or medication interactions. Conversely, unnecessary neighbors can introduce noise, increase computation, and dilute salient clinical signals. Therefore, personalized concept representation requires a graph-budgeting mechanism that adaptively determines how much relational context to retrieve for each observed code within a patient's history, rather than committing to a single global expansion budget.

The second challenge concerns the semantic text aspect of the TKG: how to use LLMs as patient-personalized semantic refiners rather than generic text encoders. Although LLMs can encode concept semantics, isolated encoding is insufficient because a code's meaning depends on co-occurring codes, visit position, temporal history, and selected KG context. We therefore define refinement as adapting a code's general semantics to the patient's specific clinical context, enabling patient-specific representations of the same code. To refine semantic knowledge in interaction with patient-specific EHR and TKG structure, the key question is how to expose such structure to the LLM. Directly serializing the full patient history and all KG neighbors into text is inefficient, redundant, and does not explicitly preserve the relational scaffold. Thus, patient-personalized semantic refinement requires a compact mechanism that lets the LLM jointly reason over textual semantics, longitudinal EHR context, and selected relational evidence.

In this work, we propose a medical concept encoder for EHR prediction named \textbf{\model{}}: KG-Aware LLM Refiner for Budgeted Patient-Personalized Concept Encoding. Starting from a global TKG over EHR codes, \model{} constructs a personalized graph for each patient's EHR history. For every observed code, a sequential reinforcement learning policy selects a code-specific KG expansion budget, allowing the same medical code to receive rich multi-hop context for one patient but minimal context for another. The policy makes these decisions sequentially, accounting for interactions among observed codes and the cumulative relational context already added to the patient graph. The resulting graph is encoded by two parallel branches: a heterogeneous GNN captures relation-aware structural dependencies, while a frozen LLM refiner uses graph-aware mixed prompts to refine code semantics through interactions among hard text tokens, code soft tokens, and KG-context soft tokens. The hard tokens describe the task and visit-code structure, while the soft tokens compactly represent observed codes and their selected KG context. Finally, the two representations are fused into patient-personalized medical concept embeddings for downstream EHR prediction. Our contributions are:
\vspace{-2mm}
\begin{itemize}[leftmargin=*]
    \item We introduce \model{}, a patient-personalized medical concept encoder that learns context-dependent representations for EHR codes by adapting general TKG knowledge to each patient's clinical context.
    \vspace{-2mm}

    \item We develop a patient-specific graph construction and budgeting framework that extracts patient graphs from a global TKG and uses sequential reinforcement learning to select code- and patient-specific relational context.
    \vspace{-2mm}

    \item We propose a graph-aware LLM refinement mechanism that encodes observed codes and selected KG neighborhoods as soft prompt tokens, enabling semantic refinement through interactions with the patient-specific visit-code scaffold.
    \vspace{-6mm}

    \item We integrate the resulting personalized concept representations into standard EHR prediction backbones and evaluate their effectiveness on sequential diagnosis prediction.

\end{itemize}

\section{Methodology}
\label{sec:methodology}

\begin{figure*}[t]
    \centering
    \includegraphics[width=0.95\textwidth]{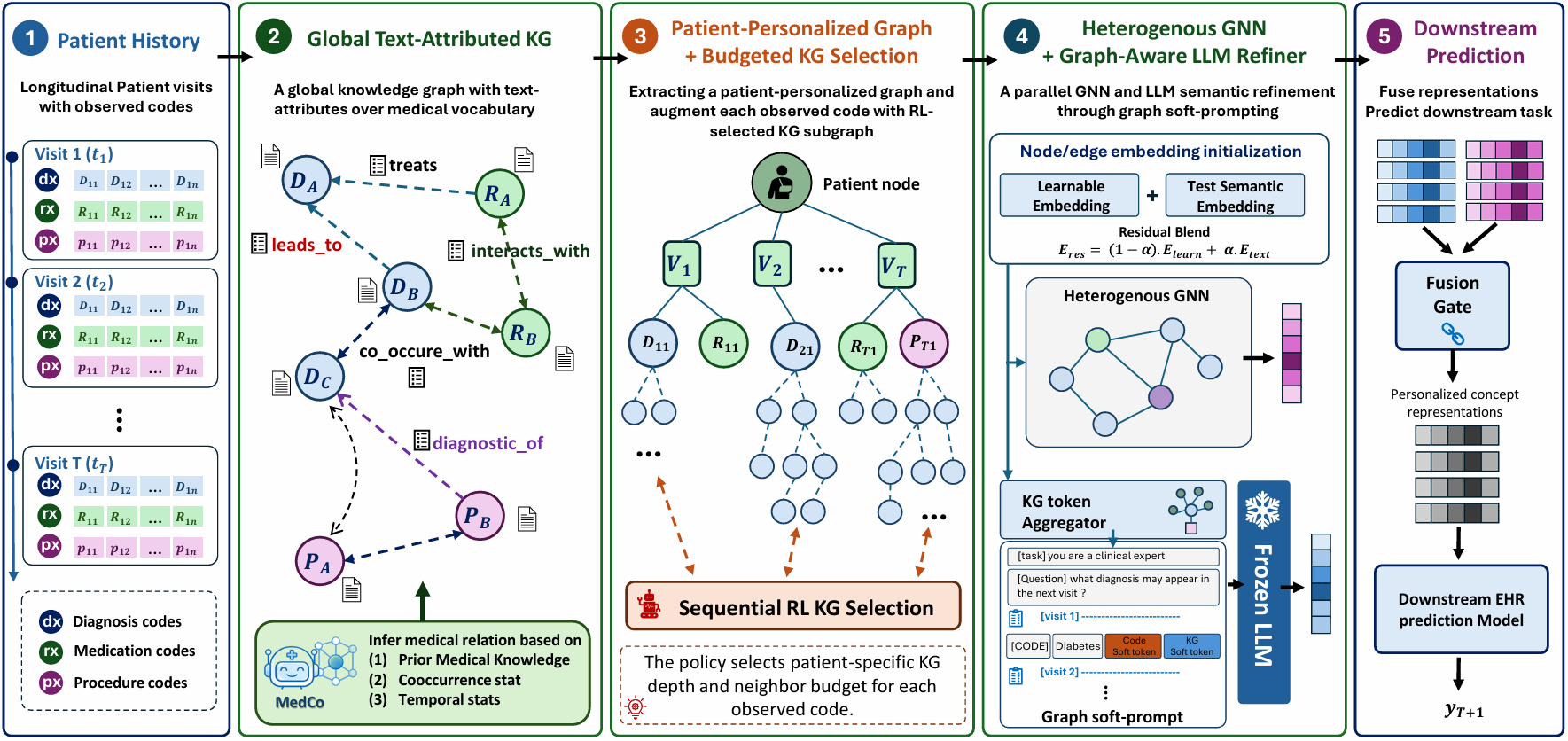} % or .png
    \caption{Overview of \model{}. Starting from longitudinal EHR visits and a global text-attributed medical KG, \model{} constructs a patient-personalized graph with RL-selected KG context for each observed code. A heterogeneous GNN and frozen LLM refiner process the selected graph in parallel, and their fused representations are used for downstream EHR prediction.}
    \label{fig:Model}
     \vspace{-3mm}
\end{figure*}

\subsection{Notation and Problem Definition}
\label{sec:notation_problem}

\paragraph{EHRs.}
Let $\mathcal{P}$ denote the patient set and let
$\mathcal{C}=\mathcal{C}^{\mathrm{dx}}\cup
\mathcal{C}^{\mathrm{rx}}\cup\mathcal{C}^{\mathrm{px}}$
denote the global vocabulary of diagnosis, medication, and procedure codes.
The EHR of patient $p\in\mathcal{P}$ is a temporally ordered visit sequence
$X_p=(V_{p,1},\ldots,V_{p,T_p})$, where each visit
$V_{p,t}\subseteq\mathcal{C}$ is a set of observed codes. %We denote the clinical type of code $c$ by $\tau(c)\in\mathcal{T}_c$, where $\mathcal{T}_c=\{\mathrm{dx},\mathrm{rx},\mathrm{px}\}$.

\paragraph{Predictive Task.}
Given the observed history
$X_p^{\leq t^\ast}=(V_{p,1},\ldots,V_{p,t^\ast})$, our goal is to learn a
patient-personalized medical concept encoder. For each observed code
$c\in V_{p,t}$, $1\le t\le t^\ast$, the encoder $\Phi_{\Theta}$ produces
\begin{equation}
    \mathbf{z}_{p,t,c}
    =
    \Phi_{\Theta}
    \left(
    c,
    X_p^{\leq t^\ast}
    \right)
    \in
    \mathbb{R}^{d},
\end{equation}

The resulting representations are used by a downstream EHR prediction model.
For next-visit diagnosis prediction, we predict:
\begin{equation}
\hat{\mathbf y}_p
=
f_\Omega\!\left(
\left\{
\operatorname{Pool}\!\left(
\{z_{p,t,c}:c\in V_{p,t}\}
\right)
\right\}_{t=1}^{t^\ast}
\right),
\end{equation}
where $\hat{\mathbf y}_p$ is next-visit label and $\hat{\mathbf y}_p \in [0,1]^{|\mathcal C^{\mathrm{dx}}|}$.

\subsection{Method Summary}
\label{sec:method_summary}

We propose \model{}, a patient-personalized medical concept encoder with five steps, shown in Figure~\ref{fig:Model}. (1) We construct a global text-attributed KG whose nodes are medical codes, with LLM-inferred semantic relations. (2) For each patient, we extract a patient-specific graph containing patient, visit, observed-code, and selected KG-neighbor nodes, connected by structural EHR edges and retrieved KG edges. (3) \model{} learns a graph-budgeting policy that selects how much KG context to retrieve for each observed code, allowing expansion depth and neighbor budget to vary for codes across patients. (4) The budgeted patient graph is processed by parallel GNN and LLM branches: the GNN captures structural dependencies through message passing, while the LLM refines concept representations based on patient semantic context through mixed prompting. (5) \model{} fuses the GNN and LLM outputs into patient-personalized concept embeddings for downstream EHR prediction.

\subsection{Global TKG Construction}
\label{sec:global_tkg_construction}
We first construct a global TKG over the full medical code vocabulary following \citet{kerdabadi2026text}. Candidate code pairs are identified from longitudinal EHRs using statistically reliable intra-visit co-occurrences and next-visit transitions. For each pair, we compute support, conditional probabilities, smoothed PMI, and statistical significance, and filter non-significant pairs. The remaining candidate pairs are passed to a type-constrained LLM relation induction module. Given two codes, EHR-derived statistics, and a type-specific relation set, the LLM assigns a directed semantic relation and a short clinical rationale, yielding clinically meaningful relations such as \texttt{treats}, \texttt{leads\_to}, and \texttt{diagnostic\_of}. Nodes and edges are further associated with textual attributes, including concept descriptions and relation rationales. Appendix~\ref{app:global_tkg_construction} provides details of the global TKG construction. We denote the resulting TKG as:
\begin{equation}
    \mathcal{G}
    =
    \left(
    \mathcal{V},
    \mathcal{E},
    \phi,
    \psi
    \right),
\end{equation}
where $\mathcal{V}=\mathcal{C}$ is the set of medical concept nodes and
$\mathcal{E}\subseteq\mathcal{V}\times\mathcal{V}$ is the set of directed edges.
The mapping $\phi:\mathcal{V}\rightarrow\mathcal{T}$ assigns each node to a
clinical type, and $\psi:\mathcal{E}\rightarrow\mathcal{R}$ assigns each edge to
a semantic relation type. For an edge $e=(u,v)\in\mathcal{E}$, we write
$\psi(e)=r$ and equivalently denote the typed relation as $(u,r,v)$.
Each node $v$ is associated with a textual concept description $d_v$, and each
edge $e$ is associated with a textual relation rationale $\eta_e$. 
Thus, $\mathcal{G}$ is a heterogeneous TKG used as the semantic substrate for extracting patient-specific graphs. Unlike global concept encoders that assign a single embedding to each code, REFINE produces instance-specific code representations conditioned on patient history and the local KG neighborhood.

\subsection{Patient-Personalized Graph Construction}
\label{sec:patient_tkg_construction}

Given patient $p$'s history
$X_p^{\leq t^\ast}=\{V_{p,t}\}_{t=1}^{t^\ast}$, we construct a
patient-personalized graph
\begin{equation}
    \mathcal{G}_p^{\leq t^\ast}
    =
    \left(
    \mathcal{V}_p,
    \mathcal{E}_p
    \right).
\end{equation}
Let
$\mathcal{S}_p=\bigcup_{t=1}^{t^\ast}V_{p,t}$
denote the set of distinct medical codes observed in the patient history. For each distinct observed code $c\in S_p$, REFINE selects a patient-dependent KG subgraph from $\mathcal G$:
\begin{equation}
    \mathcal{G}_{p,c}^{\mathrm{kg}}
    =
    \left(
    \mathcal{V}_{p,c}^{\mathrm{kg}},
    \mathcal{E}_{p,c}^{\mathrm{kg}}
    \right),
    \quad
    \mathcal{G}_{p,c}^{\mathrm{kg}}\subseteq \mathcal{G}.
\end{equation}
Thus, the same medical code may receive different relational context across
different patients.

We use $u_p$ for the patient node, $u_{p,t}^{V}$ for the visit node at time
$t$, and $u_{p,c}$ for the local node associated with code $c$. The selected
KG code set is 
$\mathcal{C}^{\mathrm{kg}}_p
=
\bigcup_{c\in S_p}
\mathcal{V}^{\mathrm{kg}}_{p,c}$. The local code vocabulary is then
    $\mathcal{C}_p^{\mathrm{loc}}
    =
    \mathcal{S}_p
    \cup
    \mathcal{C}_p^{\mathrm{kg}}$. 
    Accordingly, the patient-graph node set is
\begin{equation}
    \mathcal{V}_p
    =
    \{u_p\}
    \cup
    \{u_{p,t}^{V}\}_{t=1}^{t^\ast}
    \cup
    \{u_{p,c}:c\in\mathcal{C}_p^{\mathrm{loc}}\}.
\end{equation}
The edge set contains structural EHR edges and selected KG edges:
\begin{equation}
    \mathcal{E}_p
    =
    \mathcal{E}_p^{\mathrm{ehr}}
    \cup
    \mathcal{E}_p^{\mathrm{kg}} .
\end{equation}
The structural edges connect the patient node to all visit nodes and each visit
node to its observed codes:
$\mathcal{E}_p^{\mathrm{ehr}}
=
\mathcal{E}_p^{p\leftrightarrow v}
\cup
\mathcal{E}_p^{v\leftrightarrow c}$.
Here, $\mathcal{E}_p^{p\leftrightarrow v}$ contains edges between
$u_p$ and $u_{p,t}^{V}$, and $\mathcal{E}_p^{v\leftrightarrow c}$ contains edges between $u_{p,t}^{V}$ and $u_{p,c}$ for all $c\in V_{p,t}$. 
The selected KG edges are obtained by mapping selected global KG edges to their
corresponding local code nodes:
\begin{equation}
    \mathcal{E}_p^{\mathrm{kg}}
    =
    \left\{
    (u_{p,a},u_{p,b})
    :
    (a,b)\in
    \bigcup_{c\in\mathcal{S}_p}
    \mathcal{E}_{p,c}^{\mathrm{kg}}
    \right\}.
\end{equation}
% The selected KG edges are obtained by mapping global KG edges to their
% corresponding local nodes. Let
% \begin{equation}
%     \mathcal{E}_{p}^{\mathrm{kg,glob}}
%     =
%     \bigcup_{t=1}^{t^\ast}
%     \bigcup_{c\in V_{p,t}}
%     \mathcal{E}_{p,t,c}^{\mathrm{kg}} .
% \end{equation}
% Then
% \begin{equation}
%     \mathcal{E}_p^{\mathrm{kg}}
%     =
%     \left\{
%     (u_{p,a},u_{p,b})
%     :
%     (a,b)\in
%     \mathcal{E}_{p}^{\mathrm{kg,glob}}
%     \right\}.
% \end{equation}
Each selected KG edge inherits its semantic relation type and textual rationale
from the global TKG $\mathcal{G}$. The neighborhood selection policy is
described in Sec.~\ref{sec:rl_subgraph_extraction}, determining how
much relational context to retrieve for each observed code.

\begin{figure}[t]
    \centering
    \includegraphics[width=0.48\textwidth]{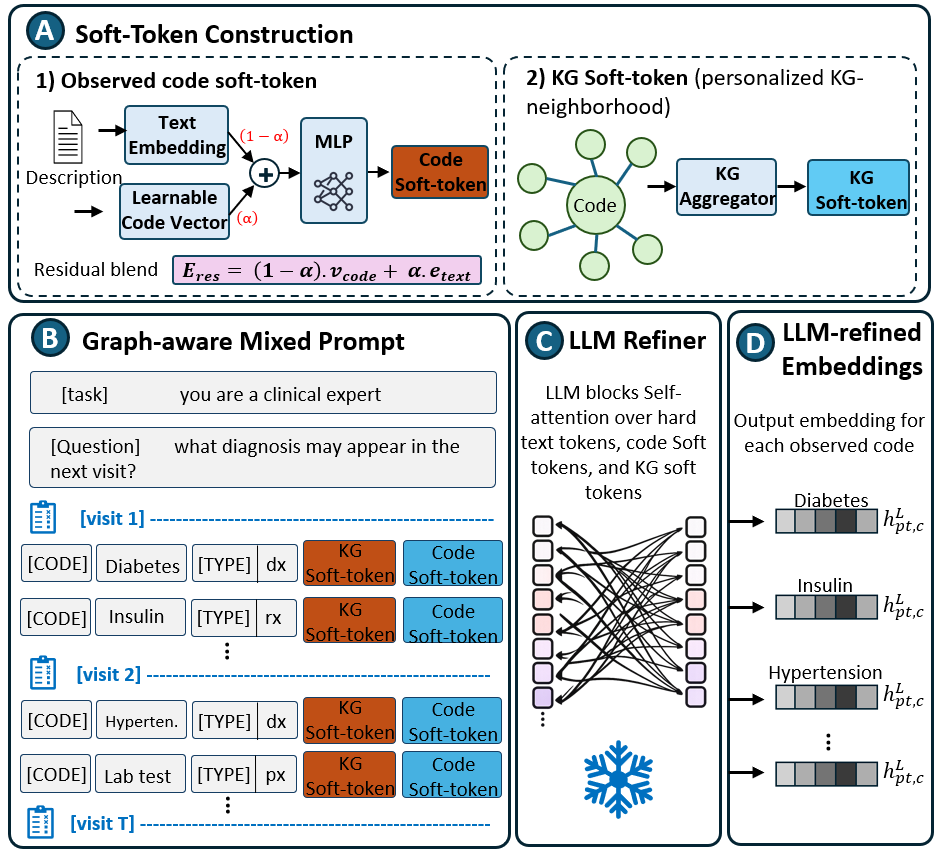} % or .png
    \caption{Graph-aware mixed prompting. Patient visits are serialized into a structured prompt with hard text tokens, code soft tokens, and KG-context soft tokens, enabling a frozen LLM to generate personalized semantic representations for observed medical codes.}
    \label{fig:llm}
     \vspace{-3mm}
\end{figure}

\subsection{Personalized Concept Sub-KG Extraction}
\label{sec:rl_subgraph_extraction}
Using the same KG expansion budget for all codes is suboptimal, since the utility of relational context depends on the patient history and code ambiguity. We therefore learn a patient-specific policy that decides how much KG context to retrieve for each distinct observed code root in the patient history. For each root code $c \in S_p$, REFINE selects a patient-specific KG subgraph from the global TKG.

\paragraph{Expansion actions.}
For each distinct observed code root $c \in S_p$, we define a discrete expansion action
$a=(K,\mathbf{m}) \in \mathcal{A}$, where $K \in \{0,\ldots,K_{\max}\}$ is the KG expansion depth and
$\mathbf{m}=(m_1,\ldots,m_K)$ specifies the per-hop fanout. The no-expansion action is
$a_0=(0,\emptyset)$. For example, $(2,(2,1))$ expands up to two hops from the observed code:
it first retains the top-2 one-hop neighbors and then the top-1 second-hop neighbor for each selected
first-hop node. Neighbors are not sampled randomly; candidate edges are ranked using evidence scores
derived from the statistical evidence computed during global KG construction. Thus, the RL policy
controls the patient-code-specific expansion budget, while neighbor selection within each budget remains
evidence-grounded.

\paragraph{Sequential decision process.}
The policy makes sequential decisions over the distinct observed code roots $S_p$. Initially, each root code is assigned the no-expansion action $a_0$. At step $\ell$, the policy selects a macro-action
$b_\ell=(c,a)$, which updates the expansion action for one distinct observed code root.
This sequential formulation allows each decision to condition on the KG
context already selected for other codes. Let $\mathcal{E}_{p}^{\mathrm{kg},(\ell)}$ and
$\mathcal{C}_{p}^{\mathrm{kg},(\ell)}$ denote the selected global KG edges and
selected KG codes after step $\ell$. The incremental cost of assigning action $a$ to root code $c$ is
\begin{equation}
\begin{aligned}
    \Delta C_{\ell}(c,a)
    =
    &~\omega_E
    \left|
    \mathcal{E}_{p,c}^{\mathrm{kg}}(a)
    \setminus
    \mathcal{E}_{p}^{\mathrm{kg},(\ell)}
    \right|
    \\
    &+
    \omega_C
    \left|
    \mathcal{V}^{\mathrm{kg}}_{p,c}(a)
    \setminus
    \left(
    \mathcal{C}^{\mathrm{kg},(\ell)}_p \cup S_p
    \right)
\right|
\end{aligned}
\end{equation}
The process terminates when the policy selects a stop action, reaches the
maximum number of steps, or exhausts the graph budget.

\paragraph{Policy and value networks.}
At step $\ell$, the policy observes a state vector $\mathbf{s}_{\ell}$ that
summarizes the patient history, selected KG context, and remaining budget. For
each feasible macro-action $b=(c,a)$, we construct an action feature
$\boldsymbol{\xi}_{\ell}(b)$ containing the state, code representation, action
embedding, and incremental cost. The actor scores feasible macro-actions as
\begin{equation}
    \pi_{\eta}(b\mid \mathbf{s}_{\ell})
    =
    \operatorname{softmax}_{b}
    \left(
    f_{\eta}(\boldsymbol{\xi}_{\ell}(b))
    \right),
\end{equation}
and the critic estimates the state value
\begin{equation}
    V_{\omega}(\mathbf{s}_{\ell})
    =
    f_{\omega}(\mathbf{s}_{\ell}).
\end{equation}
During training, macro-actions are sampled from $\pi_{\eta}$ for exploration;
during validation and inference, we use greedy selection.

\paragraph{Reward.}
During Stage I, after the policy constructs the selected patient graph
$\mathcal G_p^{\mathrm{sel}}$, we evaluate it using the frozen GNN-only
proxy predictor. Let $\mathcal L_p(\mathcal G)$ denote the corresponding
prediction loss for patient $p$ under graph $\mathcal G$, and let $\mathcal{G}_{p}^{0}$ be the baseline graph
obtained by assigning $a_0$ to all observed codes. The terminal reward
is
\begin{equation}
    R_p
    =
    \mathcal{L}_{p}(\mathcal{G}_{p}^{0})
    -
    \mathcal{L}_{p}(\mathcal{G}_{p}^{\mathrm{sel}})
    -
    \lambda_{\mathrm{cost}} C_p .
\end{equation}
Here, the graph cost is
\begin{equation}
    C_p
    =
    \omega_E|\mathcal{E}_{p}^{\mathrm{kg},\mathrm{sel}}|
    +
    \omega_C|\mathcal{C}_{p}^{\mathrm{kg},\mathrm{sel}}\setminus\mathcal{S}_p|.
\end{equation}
Thus, the policy is rewarded for improving prediction relative to the
no-expansion graph while being penalized for adding unnecessary KG context.

\paragraph{RL objective.}
Let $\tau_p=\{b_\ell\}_{\ell=1}^{L_p}$ denote the selected macro-action
trajectory for patient $p$. We optimize an actor--critic objective with entropy
regularization. The advantage at step $\ell$ is
\begin{equation}
    A_{\ell}
    =
    R_p
    -
    V_{\omega}(\mathbf{s}_{\ell}).
\end{equation}
The RL loss is
% \begin{equation}
% \begin{aligned}
%     \mathcal{L}_{\mathrm{RL}}
%     =
%     &-
%     \lambda_{\mathrm{actor}}
%     \mathbb{E}_{p}
%     \sum_{\ell=1}^{L_p}
%     \log
%     \pi_{\eta}(b_\ell\mid\mathbf{s}_{\ell})
%     A_{\ell}
%     \\
%     &+
%     \lambda_{\mathrm{critic}}
%     \mathbb{E}_{p}
%     \sum_{\ell=1}^{L_p}
%     \left(
%     V_{\omega}(\mathbf{s}_{\ell})-R_p
%     \right)^2
%     \\
%     &-
%     \lambda_{\mathrm{ent}}
%     \mathbb{E}_{p}
%     \sum_{\ell=1}^{L_p}
%     H\!\left(
%     \pi_{\eta}(\cdot\mid\mathbf{s}_{\ell})
%     \right).
% \end{aligned}
% \end{equation}
\begin{equation}
\small
\begin{aligned}
    \mathcal{L}_{\mathrm{RL}}
    =
    \mathbb{E}_{p}
    \sum_{\ell=1}^{L_p}
    \Big[
    &-\lambda_{act}
    \log \pi_{\eta}(b_\ell\mid\mathbf{s}_{\ell}) A_{\ell}
    \\
    &\hspace{-15mm}
    +
    \lambda_{crit}
    (V_{\omega}(\mathbf{s}_{\ell})-R_p)^2
    -
    \lambda_{ent}
    H(\pi_{\eta}(\cdot\mid\mathbf{s}_{\ell}))
    \Big].
\end{aligned}
\end{equation}
%In practice, we train the graph-selection policy in a GNN-only stage to avoid expensive LLM rollouts. The selected patient graphs are then exported and reused by the full \model{} model with frozen-LLM refinement.

\subsection{Graph-Aware LLM Refinement for Personalized Concept Representation}
\label{sec:ehr_graph_prompting}

After constructing the patient-personalized graph $\mathcal{G}_p^{\leq t^\ast}$, \model{} processes it with two parallel branches: a heterogeneous GNN branch and an LLM refinement branch. Both are grounded in the same fixed text-derived semantics from node descriptions $d_c$ and edge rationales $\eta_e$, but each branch maintains its own trainable parameters.

\paragraph{Node and edge embedding initialization.}
For each branch $q\in\{G,L\}$, where $G$ and $L$ denote the GNN and LLM
branches, respectively, we initialize code and KG-edge representations with a
residual blend of learnable embeddings and fixed text-derived semantics. For a
code node $u_{p,c}$ with description encoding $\tilde{\mathbf{d}}_c$, we define
\begin{equation}
    \mathbf{h}_{u_{p,c}}^{q,0}
    =
    (1-\lambda_c^q)\mathbf{a}_c^q
    +
    \lambda_c^q \mathbf{W}_{d}^{q}\tilde{\mathbf{d}}_c,
\end{equation}
where $\mathbf{a}_c^q$ is a learnable code embedding. Patient and visit nodes
use learnable type and position embeddings. For a selected KG edge $e=(u_{p,a},u_{p,b})$ corresponding to the global edge
$(a,b)\in\mathcal{E}$, with relation $r_e=\psi((a,b))$ and rationale encoding
$\tilde{\boldsymbol{\eta}}_{(a,b)}$, we define
\begin{equation}
    \mathbf{e}_{e}^{q}
    =
    (1-\lambda_r^q)\mathbf{a}_{r_e}^q
    +
    \lambda_r^q \mathbf{W}_{\eta}^{q}\tilde{\boldsymbol{\eta}}_{(a,b)}.
\end{equation}
This initialization injects clinical semantics while preserving trainable
branch-specific flexibility.

\paragraph{Heterogeneous GNN branch.}
The GNN branch performs message passing over the selected patient graph $\mathcal{G}_p^{\leq t^\ast}$. Let $\chi(u)$ denote the node type of $u$ and $r_e$ denote the edge type of $e\in\mathcal{E}_p$. For each layer $\ell$, an edge-specific message is computed as
\begin{equation}
    \mathbf{m}_{u\rightarrow v}^{(\ell)}
    =
    f_{r_e}^{\mathrm{msg}}
    \left(
    \mathbf{h}_{u}^{G,\ell},
    \mathbf{e}_{e}^{G}
    \right),
    \qquad e=(u,v).
\end{equation}
Messages are aggregated by relation type and used to update the destination node with a type-specific update function:
\begin{equation}
    \mathbf{m}_{v,r}^{(\ell)}
    =
    \operatorname{AGG}_{r}
    \{\mathbf{m}_{u\rightarrow v}^{(\ell)}:
    e\in\mathcal{E}_p,\ r_e=r\},
\end{equation}
\begin{equation}
    \mathbf{h}_{v}^{G,\ell+1}
    =
    f_{\chi(v)}^{\mathrm{upd}}
    \left(
    \mathbf{h}_{v}^{G,\ell},
    \{\mathbf{m}_{v,r}^{(\ell)}\}_{r\in\mathcal{R}_p(v)}
    \right).
\end{equation}
where
$\mathcal{R}_p(v)=\{r_e:e=(u,v)\in\mathcal E_p\}$
denotes the set of incoming edge types at node $v$. After $L$ layers, the GNN produces a structural representation
$h^G_{p,c}\in\mathbb{R}^d$ for each distinct observed code, refined by the patient-specific selected graph.

\paragraph{LLM refiner branch.}
For the LLM branch, each observed code is represented by a code soft token and a KG-context soft token. The code soft token is obtained by projecting the LLM-branch residual code representation into the LLM hidden space:
    $\mathbf{s}_{p,c}
    =
    \mathbf{W}_{\mathrm{soft}}^{c}\mathbf{h}_{u_{p,c}}^{L,0}
    \in\mathbb{R}^{d_{\mathrm{LM}}}$.
To summarize the selected KG context $\mathcal{G}_{p,c}^{\mathrm{kg}}$, we aggregate the LLM-branch representations of its selected neighbor nodes and relations:
\begin{equation}
\begin{aligned}
    \mathbf{k}_{p, c}
    =
    \operatorname{KGPool}
    \Big(
    &\{\mathbf{h}_{u_{p,v}}^{L,0}:v\in\mathcal{V}_{p,c}^{\mathrm{kg}}\},
    \\
    &\{\mathbf{e}_{e}^{L}:e\in\mathcal{E}_{p,c}^{\mathrm{kg}}\}
    \Big).
\end{aligned}
\end{equation}

$\operatorname{KGPool}(\cdot)$ applies a lightweight
GNN-style aggregator over the selected code-specific KG subgraph, combining
selected neighbor-node embeddings and relation/rationale-aware edge embeddings, producing a compact KG-context token for code $c$.

\paragraph{Mixed Graph Prompt.}
We keep the LLM frozen and construct a mixed prompt without modifying its
tokenizer or embedding table. Hard tokens encode the task, visit structure, code names, and code types, while soft tokens inject code-level and KG-context semantics. For patient $p$, the prompt follows a visit-code hierarchy:
\begin{equation}
    \mathcal{J}(p,t^\ast)
    =
    \left[
    x_{\mathrm{task}},
    x_{\mathrm{quest}},
    \{\Gamma(V_{p,t})\}_{t=1}^{t^\ast}
    \right],
\end{equation}
where $x_{\mathrm{task}}$ and $x_{\mathrm{quest}}$ denote the hard-token
task instruction and prediction query, respectively, and each visit block contains its observed codes,
\begin{equation}
    \Gamma(V_{p,t})
    =
    \left[
    \Gamma(p,t,c): c\in V_{p,t}
    \right],
\end{equation}
and each code block is represented as
\begin{equation}
    \Gamma(p,t,c)
    =
    \left[
    \mathrm{name}(c),
    \phi(c),
    \langle \mathbf{s}_{p,c}\rangle,
    \langle \mathbf{k}_{p,c}\rangle
    \right].
\end{equation}

\paragraph{LLM refinement branch.}
Let $\mathbf{X}_p\in\mathbb{R}^{M_p\times d_{\mathrm{LM}}}$ be the final mixed embedding sequence obtained by interleaving LLM token embeddings for hard text with code and KG soft tokens. The frozen LLM encodes this sequence as
\begin{equation}
    \mathbf{H}_{p}^{L}
    =
    \operatorname{LLM}
    \left(
    \mathbf{X}_p
    \right).
\end{equation}
For each observed code $c\in V_{p,t}$, we extract the hidden state at its code-soft-token position and project it back to the model hidden space:
\begin{equation}
    \mathbf{h}_{p,t,c}^{L}
    =
    \mathbf{W}_{L}
    \mathbf{H}_{p,i(p,t,c)}^{L}
    \in\mathbb{R}^{d},
\end{equation}
where $i(p,t,c)$ denotes the prompt position of the code soft token. This representation captures semantic interactions among hard textual context, code soft tokens, and selected KG-context tokens. When a code appears in multiple visits, its input code and KG-context soft tokens are shared across occurrences, while the resulting LLM hidden states remain visit-specific through their distinct prompt positions.

\paragraph{Fusion of GNN and LLM Representations.}
The GNN and LLM branches capture complementary patient-personalized information: $\mathbf{h}_{p,c}^{G}$ represents graph-structural context, while $\mathbf{h}_{p,t,c}^{L}$ represents LLM-refined semantic context. We combine them using gated fusion:
\begin{align}
    \mathbf{g}_{p,t,c}
    &=
    \sigma
    \left(
    \mathbf{W}_{F}
    [\mathbf{h}_{p,c}^{G};
    \mathbf{h}_{p,t,c}^{L}]
    \right),\\
    \mathbf{z}_{p,t,c}
    &=
    \mathbf{g}_{p,t,c}\odot \mathbf{h}_{p,t,c}^{L}
    +
    (1-\mathbf{g}_{p,t,c})\odot \mathbf{h}_{p,c}^{G}
\end{align}
$\mathbf{z}_{p,t,c}$ is then used for downstream EHR prediction.

\subsection{Integration with Downstream Models}
\label{sec:integrating_encoder}

\model{} serves as a plug-in concept encoder for standard EHR models. Given
the observed history $X_p^{\leq t^\ast}$, it produces a patient-personalized representation for each observed code $c\in V_{p,t}$:
\begin{equation}
    \mathbf{z}_{p,t,c}
    =
    \model{}_{\Theta}
    \left(
    c,
    X_p^{\leq t^\ast},
    \mathcal{G}
    \right)
\end{equation}
These code representations are aggregated within each visit and passed to an EHR backbone:
\begin{equation}
    \mathbf{h}_{p}
    =
    f_{\Omega}
    \left(
    \left\{
    \operatorname{Pool}
    \left(
    \{\mathbf{z}_{p,t,c}:c\in V_{p,t}\}
    \right)
    \right\}_{t=1}^{t^\ast}
    \right)
\end{equation}
The final next-visit diagnosis prediction is
\begin{equation}
    \widehat{\mathbf{y}}_{p}
    =
    \sigma
    \left(
    \mathbf{W}_{o}\mathbf{h}_{p}
    +
    \mathbf{b}_{o}
    \right),
    \qquad
    \widehat{\mathbf{y}}_{p}
    \in
    [0,1]^{|\mathcal{C}^{\mathrm{dx}}|}.
\end{equation}
In Stage II, with the selected patient graphs fixed, we train the full model using multi-label binary cross-entropy between
$\widehat{\mathbf{y}}_{p}$ and the next-visit diagnosis label
$\mathbf{y}_{p}$.

\begin{figure*}[t]
    \setlength{\abovecaptionskip}{3pt} % Adjust the space above the caption
    \begin{center}
    \includegraphics[width=1.0\linewidth]{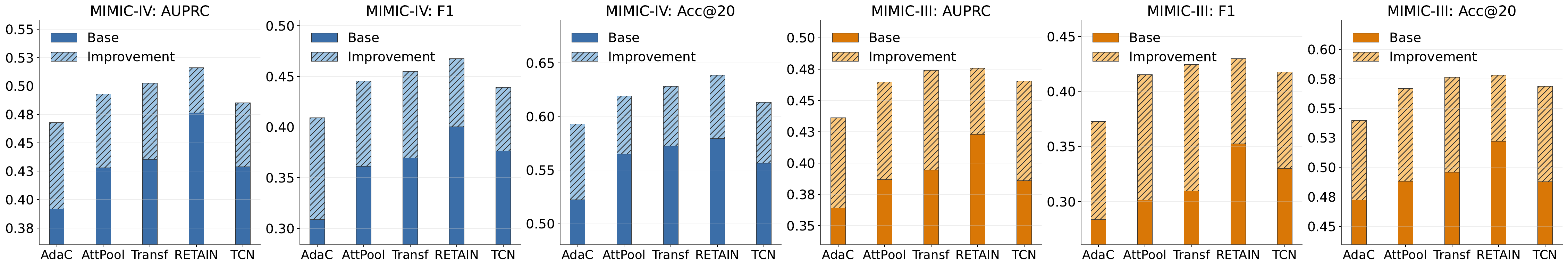}
    \caption{Performance gains from integrating \model{} as a plug-in concept encoder across prediction backbones.} %, measured by AUPRC and Acc@20.}
    \label{fig:enhancement}
    \end{center}
    \vspace{-3mm}
\end{figure*}

\begin{table*}[t]
    \centering
    \scriptsize
    \setlength{\tabcolsep}{4.2pt}
    \renewcommand{\arraystretch}{1.08}
    \caption{Baseline performance comparison. We report overall AUPRC, F1, and Acc@k, and stratified AUPRC by label frequency quartiles. The first row corresponds to the selected base model, Transformer, and subsequent rows denote plug-in variants, for example \model{} = \model{} + Base. All values are reported as percentages.}
    \vspace{-1mm}
    \begin{tabular}{
        c l
        *{5}{S[table-format=2.2]}
        *{4}{S[table-format=2.2]}
    }
        \toprule
        & \multirow{2}{*}{\textbf{Model}} &
        \multicolumn{5}{c}{\textbf{General Performance}} &
        \multicolumn{4}{c}{\textbf{Label-Frequency AUPRC}} \\
        \cmidrule(lr){3-7} \cmidrule(lr){8-11}
        & &
        {\textbf{AUPRC}} & {\textbf{F1}} & {\textbf{Acc@15}} & {\textbf{Acc@20}} & {\textbf{Acc@30}} &
        {\textbf{0--25\%}} & {\textbf{25--50\%}} & {\textbf{50--75\%}} & {\textbf{75--100\%}} \\
        \midrule
        \multirow{13}{*}{\rotatebox{90}{\textbf{MIMIC-III}}}
        & Base~\citep{vaswani2017attention}  & 39.44 & 30.96 & 46.47 & 49.60 & 57.73 & 38.64 & 47.30 & 70.63 & 74.30 \\
        & GRAM~\citep{choi2017gram}  & 40.10 & 32.32 & 47.75 & 51.02 & 58.71 & 40.06 & 49.84 & 71.78 & 75.16 \\
        & MMORE~\citep{song2019medical} & 40.95 & 33.08 & 48.42 & 51.73 & 59.34 & 40.72 & 50.92 & 72.55 & 75.86 \\
        & KAME~\citep{ma2018kame}  & 40.54 & 32.83 & 48.17 & 51.20 & 58.99 & 39.70 & 49.49 & 71.70 & 75.02 \\
        & G-BERT~\citep{shang2019pre}& 40.83 & 33.15 & 48.47 & 51.49 & 59.22 & 39.93 & 49.78 & 71.94 & 75.18 \\
        & HAP~\citep{zhang2020hierarchical}   & 40.70 & 32.95 & 48.30 & 51.34 & 59.12 & 39.82 & 49.63 & 71.82 & 75.10 \\
        & ADORE~\citep{cheong2023adaptive} & 40.92 & 33.20 & 48.55 & 51.63 & 59.28 & 40.05 & 49.86 & 71.99 & 75.21 \\
        & KAMPNet~\citep{an2023kampnet} & 41.37 & 33.70 & 49.10 & 52.20 & 59.82 & 40.50 & 50.59 & 72.50 & 75.78 \\
        & GraphCare~\citep{jiang2023graphcare} & 43.35 & 35.46 & 52.76 & 56.00 & 62.75 & 44.80 & 58.16 & 70.97 & 65.72 \\
        & Rel-LLM~\citep{wu2025large} & 43.58 & 36.05 & 52.90 & 56.08 & 62.92 & 44.95 & 58.35 & 73.85 & 78.40 \\
        & LINKO~\citep{nayebi2025multi} & 44.10 & 37.05 & 53.05 & 56.20 & 63.00 & 45.20 & 59.05 & \underline{74.25} & 79.20 \\
        & MedCo~\citep{kerdabadi2026text} & \underline{45.35} & \underline{39.90} & \underline{53.22} & \underline{56.32} & \underline{63.18}
        & \underline{45.73} & \underline{64.00} & 73.80 & \underline{82.80} \\
        \rowcolor{gray!10}
        & \textbf{\model{}} & \textbf{47.41} & \textbf{42.46} & \textbf{54.38} & \textbf{57.62} & \textbf{64.60}
        & \textbf{47.92} & \textbf{66.55} & \textbf{76.28} & \textbf{86.95} \\
        \addlinespace[1mm]
        
        \multirow{13}{*}{\rotatebox{90}{\textbf{MIMIC-IV}}}
        & Base~\citep{vaswani2017attention}  & 43.56 & 36.94 & 54.02 & 57.24 & 63.64 & 44.93 & 53.50 & 51.85 & 73.10 \\
        & GRAM~\citep{choi2017gram}  & 44.72 & 37.78 & 54.76 & 58.02 & 64.56 & 45.76 & 54.39 & 52.60 & 73.80 \\
        & MMORE~\citep{song2019medical} & 45.28 & 38.35 & 55.27 & 58.54 & 64.96 & 46.25 & 54.88 & 53.47 & 74.30 \\
        & KAME~\citep{ma2018kame}  & 44.39 & 37.43 & 54.45 & 57.65 & 64.09 & 45.22 & 54.05 & 52.18 & 73.52 \\
        & G-BERT~\citep{shang2019pre}& 44.80 & 37.86 & 54.86 & 58.06 & 64.50 & 45.63 & 54.52 & 52.76 & 73.94 \\
        & HAP~\citep{zhang2020hierarchical}   & 44.75 & 37.70 & 54.75 & 58.00 & 64.43 & 45.55 & 54.37 & 52.69 & 73.84 \\
        & ADORE~\citep{cheong2023adaptive} & 44.94 & 38.00 & 55.00 & 58.19 & 64.65 & 45.78 & 54.64 & 52.90 & 74.02 \\
        & KAMPNet~\citep{an2023kampnet} & 45.48 & 38.65 & 55.61 & 58.86 & 65.32 & 46.31 & 55.34 & 54.00 & 74.60 \\
        & GraphCare~\citep{jiang2023graphcare} & 46.90 & 40.12 & 55.89 & 59.14 & 65.71 & 48.31 & \textbf{60.31} & \textbf{73.70} & 66.33 \\
        & Rel-LLM~\citep{wu2025large} & 47.20 & 40.55 & 56.45 & 59.85 & 66.35 & 48.70 & 56.80 & 56.40 & 76.60 \\
        & LINKO~\citep{nayebi2025multi} & 47.55 & 41.10 & 57.05 & 60.55 & 67.05 & 49.10 & 57.10 & 56.95 & 77.20 \\
        & MedCo~\citep{kerdabadi2026text} & \underline{48.28} & \underline{43.10} & \underline{58.30} & \underline{61.55} & \underline{67.60}
        & \underline{49.95} & 57.35 & 57.70 & \underline{78.20} \\
        \rowcolor{gray!10}
        & \textbf{\model{}} & \textbf{50.24} & \textbf{45.48} & \textbf{59.48} & \textbf{62.82} & \textbf{69.00}
        & \textbf{52.30} & \underline{57.98} & \underline{59.65} & \textbf{81.55} \\
        \bottomrule
    \end{tabular}
    \vspace{-0.2cm}
    \label{tab:baseline_comparison_w_rare_codes}
\end{table*}

\begin{table}[t]
\centering
\caption{Data statistics for MIMIC-III and MIMIC-IV.}
\label{tab:Data-Statistics}
\vspace{-2mm}
\scriptsize
\setlength{\tabcolsep}{4pt}
\renewcommand{\arraystretch}{1.00}
\begin{tabular}{lcc}
\toprule
\textbf{Metric} & \textbf{MIMIC-III} & \textbf{MIMIC-IV} \\
\midrule
\# Patients                 & 7,515   & 18,829 \\
\# Visits (samples)         & 12,430  & 25,028 \\
\# Labels/sample            & 12.31   & 10.56  \\
\# Unique conditions (ICD)  & 515     & 562    \\
\# Conditions/sample        & 22.94   & 59.50  \\
\# Drugs/sample             & 54.95   & 118.16 \\
\# Unique drugs             & 468     & 510    \\
\# Procedures/sample        & 5.92    & 5.41   \\
\# Unique procedures        & 280     & 322    \\
\bottomrule
\end{tabular}
\vspace{-5mm}
\end{table}

\begin{table}[t]
\centering
\scriptsize
\setlength{\tabcolsep}{3pt}
\renewcommand{\arraystretch}{1.12}
\caption{Component-wise ablation of \model{} on MIMIC-IV and MIMIC-III.}
\label{tab:component_ablation}
\vspace{-1mm}
\begin{tabular}{c l ccc}
\toprule
\textbf{} & \textbf{Variant} & \textbf{AUPRC} & \textbf{F1} & \textbf{Acc@20} \\
\midrule

\multirow{7}{*}{\rotatebox{90}{\textbf{MIMIC-IV}}}
& \cellcolor{gray!10}\textbf{Full \model{}}
  & \cellcolor{gray!10}\textbf{50.24}
  & \cellcolor{gray!10}\textbf{45.48}
  & \cellcolor{gray!10}\textbf{62.82} \\
& w/o RL budgeting & 49.18 & 44.18 & 61.92 \\
& w/o full LLM refiner & 49.32 & 44.05 & 61.84 \\
& w/o soft prompt tokens & 49.38 & 44.12 & 61.90 \\
& w/o KG soft tokens & 49.72 & 44.78 & 62.32 \\
& w/o full personalized GNN & 47.35 & 40.82 & 60.25 \\
& Global code representation (MedCo) & 48.28 & 43.10 & 61.55 \\

\addlinespace[0.6mm]

\multirow{7}{*}{\rotatebox{90}{\textbf{MIMIC-III}}}
& \cellcolor{gray!10}\textbf{Full \model{}}
  & \cellcolor{gray!10}\textbf{47.41}
  & \cellcolor{gray!10}\textbf{42.46}
  & \cellcolor{gray!10}\textbf{57.62} \\
& w/o RL budgeting & 46.38 & 41.02 & 56.70 \\
& w/o full LLM refiner & 46.51 & 40.88 & 56.62 \\
& w/o soft prompt tokens & 46.57 & 40.96 & 56.68 \\
& w/o KG soft tokens & 46.92 & 41.68 & 57.10 \\
& w/o full personalized GNN & 43.85 & 36.62 & 56.14 \\
& Global code representation (MedCo) & 45.35 & 39.90 & 56.32 \\

\bottomrule
\end{tabular}
\vspace{-4mm}
\end{table}

\section{Experiments}

\noindent\textbf{Experimental Setting.} We evaluate on two public EHR benchmarks, MIMIC-III \citep{johnson2016mimic} and MIMIC-IV \citep{johnson2023mimic}. Table~\ref{tab:Data-Statistics} reports cohort statistics. The task is next-visit diagnosis prediction over imbalanced label spaces (515 codes in MIMIC-III; 562 in MIMIC-IV). Implementation details, including LLM/GNN configurations and the two-stage training pipeline, are provided in Appendix~\ref{app:implementation_details}. The source code is available at \url{https://github.com/mohsen-nyb/REFINE.git}

\noindent\textbf{Evaluation Metrics.} We report results averaged over five folds for \textbf{AUPRC} (both general and also stratified by label frequency to assess performance across code rarity levels), \textbf{Acc@k} (top-$k$ accuracy normalized by $\min(k, |\mathbf{y}_{t+1}|)$), and \textbf{F1}.

\section{Results}
\vspace{-3mm}
We evaluate \model{} through five research questions:
\textbf{RQ1:} Does \model{} improve downstream EHR prediction as a plug-in concept encoder?
\textbf{RQ2:} How does \model{} compare with existing baselines?
\textbf{RQ3:} Which components contribute the most to performance?
\textbf{RQ4:} Does learned KG budgeting outperform fixed KG selection strategies? \textbf{RQ5:} Does \model{} improve prediction under data scarcity?
RQ1--RQ4 are addressed in the main results, and RQ5 in Appendix~\ref{app:appendix_data_insufficiency}.

\subsection{RQ1: Plug-in Encoder Evaluation}
\vspace{-1mm}
We evaluate whether \model{} improves standard EHR backbones when used as a
plug-in concept encoder. We integrate \model{} into five representative models:
\textbf{AttPool}, a simple attention pooling predictor; \textbf{AdaCare}~\citep{ma2020adacare};
\textbf{Transformer}~\citep{vaswani2017attention}; \textbf{RETAIN}~\citep{choi2016retain};
and \textbf{TCN}~\citep{bai2018empirical}. For each backbone, we compare the
base model with its \model{}-enhanced variant. As shown in
Figure~\ref{fig:enhancement}, \model{} consistently improves performance across
backbones, demonstrating its effectiveness as a plug-in personalized
concept encoder.

\subsection{RQ2: Baseline Comparison}
We compare \model{} with representative medical concept encoders, including the \textbf{base} Transformer~\citep{vaswani2017attention}, ontology-based methods (\textbf{GRAM}~\citep{choi2017gram}, \textbf{MMORE}~\citep{song2019medical}, \textbf{KAME}~\citep{ma2018kame}, \textbf{G-BERT}~\citep{shang2019pre}, \textbf{HAP}~\citep{zhang2020hierarchical}), and KG-based methods: \textbf{ADORE}~\citep{cheong2023adaptive} uses SNOMED relations; \textbf{GraphCare}~\citep{jiang2023graphcare} combines UMLS- and LLM-derived edges; \textbf{KAMPNet}~\citep{an2023kampnet} and \textbf{LINKO}~\citep{nayebi2025multi} construct cross-type connections; \textbf{Rel-LLM} follows relational graph prompting~\citep{wu2025large}; and \textbf{MedCo}~\citep{kerdabadi2026text} learns global KG--LLM concept embeddings from a medical TKG. All methods use the same downstream backbone. As shown in Table~\ref{tab:baseline_comparison_w_rare_codes}, \model{} achieves the best overall performance, highlighting the benefit of patient-personalized, budgeted TKG refinement. For rare-condition analysis, we stratify diagnosis labels into frequency quartiles; REFINE achieves the best AUPRC in six of eight quartile groups.

\begin{figure}[t]
    \setlength{\abovecaptionskip}{3pt} % Adjust the space above the caption
    \begin{center}
    \includegraphics[width=1.0\linewidth]{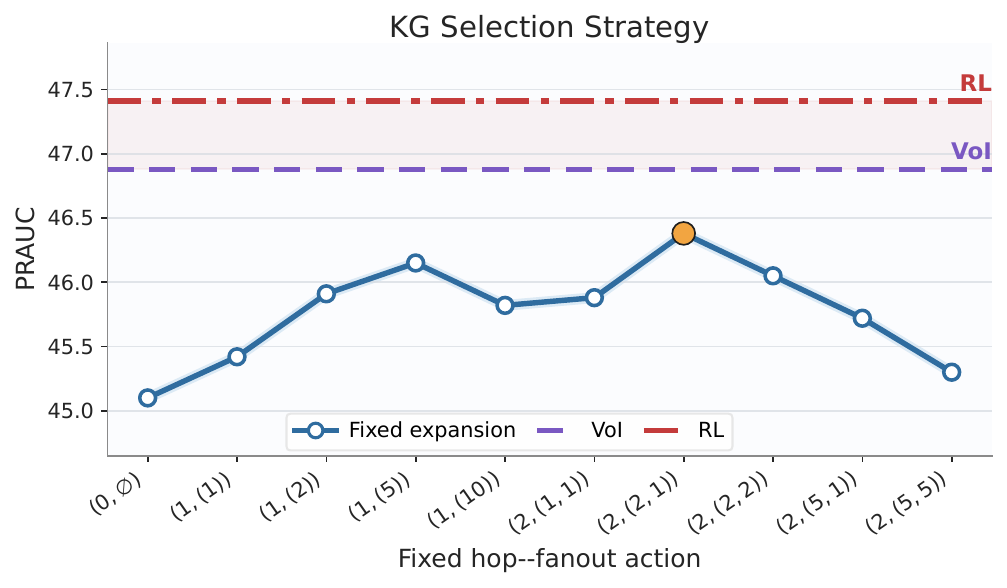}
    \caption{AUPRC comparison of KG selection strategies.}
    \label{fig:kg_selection}
    \end{center}
    \vspace{-5mm}
\end{figure}

\begin{figure}[t]
    \centering
    \includegraphics[width=\linewidth]{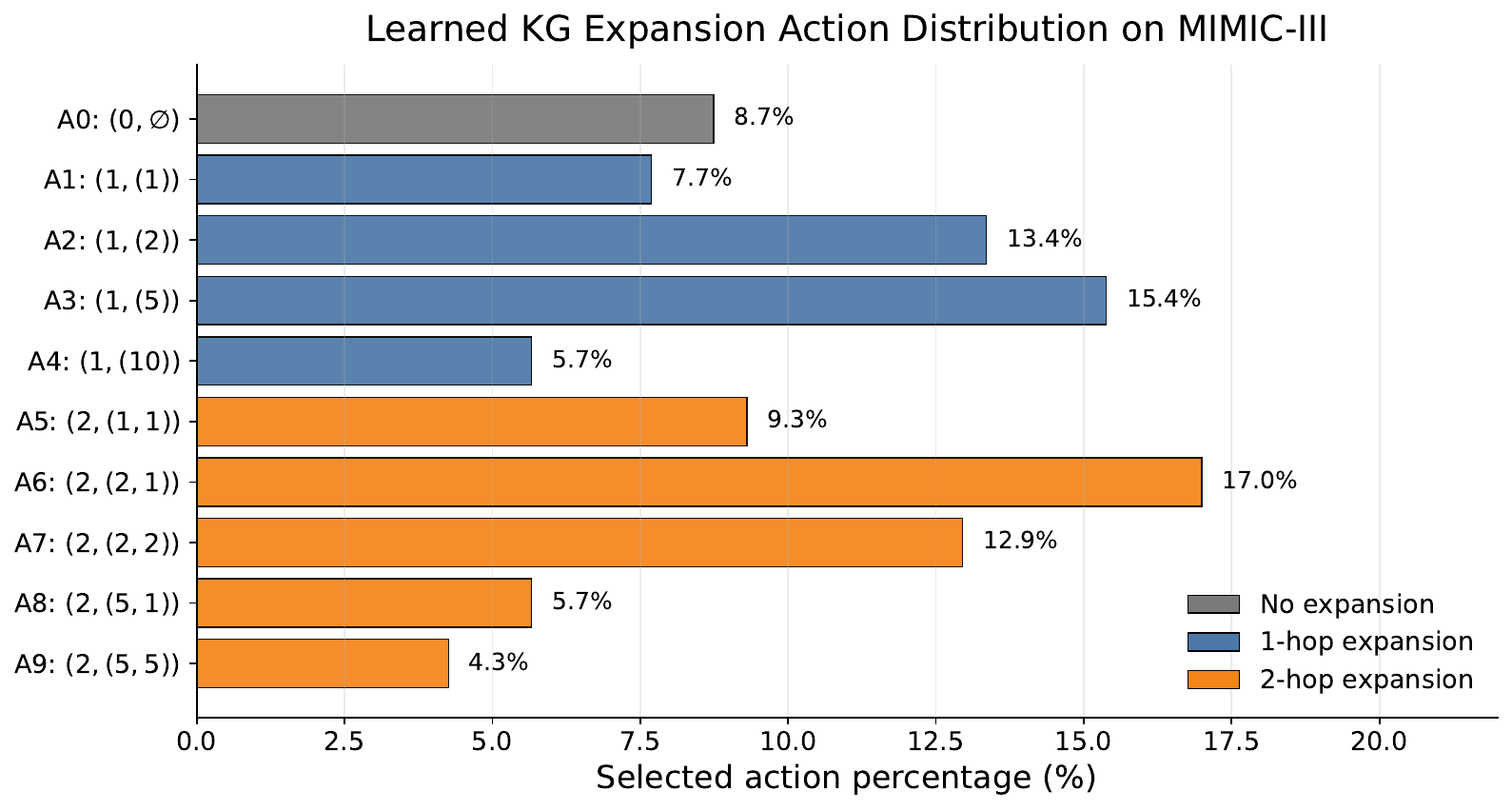}
    \caption{Distribution of learned KG expansion actions.}
    \label{fig:mimic3_kg_action_distribution}
\end{figure}

\subsection{RQ3: Ablation Study}
Table~\ref{tab:component_ablation} shows that each component contributes to the final performance of \model{}. Removing the personalized GNN yields the largest drop, confirming that the frozen LLM alone is insufficient and that patient-graph structure is central to REFINE. Removing the LLM refiner or using only hard name tokens also degrades performance, showing that semantic refinement benefits from trainable soft prompts rather than text names alone. The drop without KG soft tokens indicates that selected relational context adds value beyond observed-code soft tokens. Finally, removing RL budgeting consistently hurts performance, demonstrating the importance of adaptively selecting patient- and code-specific KG context instead of using an unbudgeted or fixed expansion. Even without the LLM refiner, \model{} outperforms MedCo, showing that personalized budgeted TKG construction alone improves over global KG-based code representations.

\begin{table*}[t]
\centering
\small
\caption{Representative examples of RL-policy budget personalization on MIMIC-III. Code frequency is the number of prediction samples containing the code at least once.}
\label{tab:RL policy_qualitative_examples}
\resizebox{\textwidth}{!}{
\begin{tabular}{p{3.0cm} r p{3.0cm} p{3.6cm} p{6.0cm}}
\toprule
\textbf{Code / Patient} &
\textbf{Code Freq.} &
\textbf{Selected Budget} &
\textbf{Patient Context} &
\textbf{Interpretation} \\
\midrule

\texttt{conditions:14.5.23}
& 7
& 2-hop $(5,1)$
& 2 visits, 9 observed codes
& A rare diagnosis receives deeper multi-hop relational context in a short patient history. \\

\texttt{procedures:7.12.2}
& 10
& 2-hop $(5,1)$
& 1 visit, 41 observed codes
& A rare procedure also receives deeper expansion, showing that adaptive budgeting is not limited to diagnosis codes. \\

\texttt{drugs:B05XA}
& 10,809
& 0-hop
& 5 visits, 211 observed codes
& A highly frequent drug receives no additional KG context, showing selective expansion. \\

\texttt{conditions:16.10.1.4}
& 683
& 1-hop $(5)$
& 5 visits, 182 observed codes
& A medium-frequency diagnosis receives only one-hop context expansion. \\

\bottomrule
\end{tabular}}
\vspace{-4mm}
\end{table*}

\subsection{RQ4: Personalized KG Selection}

Figure~\ref{fig:kg_selection} compares different KG selection strategies. Fixed strategies apply the same hop--fanout expansion to all observed codes, and their performance varies substantially across actions, indicating that a single global KG budget is suboptimal. The value-of-information (VoI) strategy\footnote{Details of the VoI strategy are provided in Appendix~\ref{app:voi_selection}.}
improves over fixed expansion by using a learned utility critic to select visit-level KG expansion actions. The critic is trained with sparse label-derived utility targets, while validation and test-time selection is label-free. In contrast, the RL policy performs sequential code-level budgeting, allowing different observed code roots to receive different KG expansions while conditioning on previously selected KG context. Its superior performance demonstrates the benefit of adaptive, fine-grained KG selection.

\paragraph{Analysis of learned RL action distribution.}
To better understand the learned selection behavior, Figure~\ref{fig:mimic3_kg_action_distribution} shows the distribution of RL policy expansion actions on MIMIC-III. The policy does not collapse to a single hop--fanout configuration, but instead selects a diverse range of no-expansion, one-hop, and two-hop actions, with moderate one-hop and two-hop budgets among the more frequently selected configurations. This diversity suggests that the selector balances predictive utility against graph-expansion cost across different patient-code instances.

\paragraph{Qualitative personalization analysis.}
To make this behavior more concrete, we inspect representative RL policy decisions from individual MIMIC-III prediction samples. Here, code frequency denotes the number of prediction samples containing a code at least once. As shown in Table~\ref{tab:RL policy_qualitative_examples}, the policy does not uniformly expand all observed code roots. For example, a rare diagnosis appearing in only 7 samples receives a 2-hop budget in a patient history with 2 visits and 9 observed codes, while a rare procedure appearing in 10 samples also receives a 2-hop budget in another patient. In contrast, a frequent medication appearing in 10,809 samples receives no KG expansion in a patient with 5 visits and 211 observed codes, while a medium-frequency diagnosis receives only local 1-hop context. 
These examples show that the RL policy assigns heterogeneous budgets across codes and patient contexts rather than uniformly expanding patient graphs.

\section{Related Work}
\label{related_work}
The increasing availability of EHRs has advanced clinical prediction methods, ranging from early sequential models~\citep{choi2016doctor,choi2016retain} to attention-based and Transformer architectures~\citep{li2020behrt,nayebi2023contrastive,hu2025recurrent,hadizadeh2024contrastive}, as well as graph-based approaches~\citep{xu2022counterfactual,yang2023molerec}. These methods primarily improve patient-level prediction architectures, whereas our focus is on learning patient-personalized representations of the medical concepts.

A major line of work improves medical code representations by incorporating pre-existing medical ontologies or KG structure. GRAM~\citep{choi2017gram}, MMORE~\citep{song2019medical}, KAME~\citep{ma2018kame}, and HAP~\citep{zhang2020hierarchical} exploit hierarchical ontology structure to refine concept embeddings through parent--child dependencies and attention mechanisms. Beyond hierarchies, G-BERT~\citep{shang2019pre}, ADORE~\citep{cheong2023adaptive}, KAMPNet~\citep{an2023kampnet}, and LINKO~\citep{nayebi2025multi} incorporate richer relational graphs, multi-source knowledge, or contrastive objectives. GCL~\citep{lu2021collaborative}, RAM-EHR~\citep{xu2024ram}, KARE~\citep{jiang2024reasoning}, and GraphCare~\citep{jiang2023graphcare} further integrate retrieved knowledge, auxiliary text, or patient-specific graph context. However, GraphCare relies on fixed neighborhood expansion, whereas \model{} learns sequential patient- and code-specific KG budgets, allowing different observed codes within the same patient to receive different amounts of relational context.

Recent work also combines LLMs with relational or medical graph representations. Rel-LLM~\citep{wu2025large} injects explicit relational structure through graph-aware soft prompting, while \model{} compresses the selected code-specific KG neighborhood into compact soft context tokens within a visit-aware prompt, avoiding explicit serialization of every retrieved relation. MedCo~\citep{kerdabadi2026text} learns a single global KG--LLM representation per medical concept. In contrast, \model{} learns patient-code representations conditioned on longitudinal history and adaptively selected KG context, using this relational evidence to guide LLM semantic refinement.

\section{Conclusion}
We presented \model{}, a patient-personalized plug-in medical concept encoder that learns EHR code representations from budgeted text-attributed patient graphs. \model{} uses sequential reinforcement learning to select code- and patient-specific KG budgets, builds compact personalized graphs, and combines a heterogeneous GNN with a frozen LLM through graph-aware soft prompts. Experiments on MIMIC-III and MIMIC-IV show consistent gains across diverse backbones and strong baselines. Ablation and KG-selection analyses confirm the complementary benefits of personalized graph construction, adaptive budgeting, and LLM refinement, while showing that the learned policy adapts budgets across codes and patient contexts. \model{} also remains effective under limited supervision and across frozen LLM backbones.

\section{Limitations and Future Work}

A limitation of our KG selection module is its predefined discrete action space with fixed hop and neighbor budgets. Although neighbors within each selected budget are chosen by evidence-based edge rankings rather than random sampling, the policy does not directly select the exact nodes or adapt the budget continuously at each hop. Future work will explore more flexible RL strategies that jointly select, rank, and budget KG nodes while remaining efficient for large-scale EHR prediction.

\section{Ethical Considerations}

We comply with the ACL Ethics Policy throughout this study. All experiments use publicly available, de-identified EHR datasets, which provide strong privacy protections. We do not send patient-level records or identifiable clinical histories to external LLM services. LLM-based KG construction uses only concept-level information and aggregated EHR-derived statistics, and the LLM used in \model{} is run locally and kept frozen. AI-assisted tools were used solely for writing and grammar checking, and not for data analysis, experimental design, or generating scientific claims. The proposed model is intended for research on EHR representation learning and should not be used for clinical decision-making without prospective validation, fairness evaluation, and expert oversight.

\section{Acknowledgments}
This research was supported by the National Science Foundation under Award No.~2544634.

\bibliography{ref}

\newpage
\appendix

\section{Appendix}

\begin{algorithm}[ht]
\footnotesize
\caption{Two-stage training of \model{}}
\label{alg:refine_training}
\begin{algorithmic}[1]
\Require Training set $\mathcal D$, global TKG $\mathcal G$, action set $\mathcal A$, 
null action $a_0$, graph budget $B_{\mathrm{kg}}$, 
minimum/maximum decision steps $L_{\min},L_{\max}$
\Ensure Trained parameters $\Theta$

\Statex \textbf{Stage I: RL-based graph budgeting}
\State Initialize and warm up GNN-only proxy predictor; freeze its parameters
\State Initialize actor--critic policy $(\pi_\eta,V_\omega)$
\For{epoch $=1,\dots,E_1$}
    \For{$p \in \mathcal D$}
        \State $\mathcal{S}_p \gets \bigcup_{t\leq t^\ast} V_{p,t}$ 
        \Comment{distinct observed code roots}
        \State $A_p \gets \{a_c \leftarrow a_0 : c\in\mathcal{S}_p\}$
        \State $\mathcal{G}^{\mathrm{sel}}_p 
        \gets \textsc{Expand}(\mathcal{S}_p,A_p,\mathcal G)$
        \For{$\ell=1,\dots,L_{\max}$}
            \If{$C_{\mathrm{kg}}(A_p)\geq B_{\mathrm{kg}}$}
                \State \textbf{break}
            \EndIf
            \State $\mathcal{U}_\ell 
            \gets \textsc{Feasible}(\mathcal{S}_p,A_p,\mathcal A,B_{\mathrm{kg}})$
            \Comment{additive, budget-feasible updates}
            \If{$\ell \geq L_{\min}$}
                \State $\mathcal{U}_\ell 
                \gets \mathcal{U}_\ell \cup \{\textsc{stop}\}$
            \EndIf
            \If{$\mathcal{U}_\ell=\emptyset$}
                \State \textbf{break}
            \EndIf
            \State Select $b_\ell \sim \pi_\eta(\cdot\mid s_\ell)$ over $\mathcal{U}_\ell$
            \If{$b_\ell=\textsc{stop}$}
                \State \textbf{break}
            \EndIf
            \State Let $b_\ell=(c_\ell,a_\ell)$; update $A_p[c_\ell]\gets a_\ell$
            \State $\mathcal{G}^{\mathrm{sel}}_p 
            \gets \textsc{Expand}(\mathcal{S}_p,A_p,\mathcal G)$
        \EndFor
        \State $R_p \gets 
        L_p(\mathcal{G}^{0}_p)
        -L_p(\mathcal{G}^{\mathrm{sel}}_p)
        -\lambda_{\mathrm{cost}}\,C_{\mathrm{kg}}(A_p)$
        \State Update $(\pi_\eta,V_\omega)$ using the actor--critic objective with reward $R_p$
    \EndFor
\EndFor
\State Export graph cache $\mathcal M$ by running the learned policy greedily for each $p\in\mathcal D$

\Statex \textbf{Stage II: Full \model{} training}
\State Initialize full \model{} with frozen LLM parameters
\For{epoch $=1,\dots,E_2$}
    \For{minibatch $\mathcal P \subset \mathcal D$}
        \For{$p \in \mathcal P$}
            \State Retrieve $\mathcal{G}^{\mathrm{sel}}_p$ from cache $\mathcal M$
            \State $H_p^G \gets \textsc{GNNEnc}(\mathcal{G}^{\mathrm{sel}}_p)$
            \State $H_p^L \gets \textsc{LLMRefine}(\mathcal{G}^{\mathrm{sel}}_p)$
            \State $Z_p \gets \textsc{Fuse}(H_p^G,H_p^L)$
            \State $\hat{y}_p \gets f_\Omega(Z_p)$
        \EndFor
        \State Update $\Theta$ using 
        $\sum_{p\in\mathcal P}\mathcal L_{\mathrm{BCE}}(\hat y_p,y_p)$
    \EndFor
\EndFor
\State \Return $\Theta$
\end{algorithmic}
\end{algorithm}

\subsection{Global Text-Attributed KG Construction}
\label{app:global_tkg_construction}
We reuse the evidence-grounded TKG construction protocol from MedCo~\cite{kerdabadi2026text}, where the induced relation schema and representative KG edges were clinically validated. Therefore, this work does not introduce a new KG induction pipeline; instead, it studies how to personalize the use of this validated global TKG for patient-specific EHR prediction.

We construct a global text-attributed knowledge graph (KG) over the medical
code vocabulary. Let
$\mathcal{C}=\mathcal{C}^{\mathrm{dx}}\cup
\mathcal{C}^{\mathrm{rx}}\cup\mathcal{C}^{\mathrm{px}}$
denote the set of diagnosis, medication, and procedure codes. The resulting
global graph is denoted as
\begin{equation}
    \mathcal{G}
    =
    \left(
    \mathcal{V},
    \mathcal{E},
    \phi,
    \psi
    \right),
\end{equation}
where $\mathcal{V}=\mathcal{C}$ is the set of medical concept nodes and
$\mathcal{E}\subseteq\mathcal{V}\times\mathcal{V}$ is the set of directed
edges. The mapping $\phi:\mathcal{V}\rightarrow\mathcal{T}_c$ assigns each
node to a clinical type, where
$\mathcal{T}_c=\{\mathrm{dx},\mathrm{rx},\mathrm{px}\}$, and
$\psi:\mathcal{E}\rightarrow\mathcal{R}$ assigns each edge to a semantic
relation type. For an edge $e=(u,v)\in\mathcal{E}$, we write
$\psi(e)=r$ and equivalently denote the typed relation as $(u,r,v)$. Each
node $v$ is associated with a textual concept description $d_v$, and each
edge $e$ is associated with a textual relation rationale $\eta_e$.

\paragraph{Candidate pair extraction.}
Each patient record is represented as a sequence of visits
$\{V_{p,1},\ldots,V_{p,T_p}\}$, where each visit contains a de-duplicated set
of medical codes. We extract candidate code pairs using two complementary
EHR-derived evidence channels. The first channel captures intra-visit
co-occurrence, where two codes appear in the same visit. The second channel
captures next-visit temporal transitions, where a source code appears in
$V_{p,t}$ and a target code appears in $V_{p,t+1}$. These two channels allow
the global KG to encode both simultaneous clinical associations and temporally
ordered dependencies.

\paragraph{Pairwise statistical evidence.}
For each candidate directed pair $(c_i,c_j)$ and evidence channel
$m\in\{\mathrm{co},\mathrm{temp}\}$, we compute support counts, a smoothed
conditional probability, a PMI-style association score, and a statistical
significance score. The smoothed conditional probability is
\begin{equation}
    P_m(c_j\mid c_i)
    =
    \frac{x_m(c_i,c_j)+\alpha}
    {x_m(c_i)+\alpha|\mathcal{C}|},
\end{equation}
where $x_m(c_i,c_j)$ is the pair count under channel $m$, $x_m(c_i)$ is the
corresponding source-code count, and $\alpha$ is a Laplace smoothing constant.
We also compute a PMI-style score:
\begin{equation}
    \mathrm{PMI}_m(c_i,c_j)
    =
    \log
    \frac{p_m(c_i,c_j)}
    {p_m^{\mathrm{src}}(c_i)p_m^{\mathrm{tgt}}(c_j)} .
\end{equation}
Here, $p_m(\cdot)$ denotes empirical probabilities estimated from the
corresponding counts. For intra-visit co-occurrence,
$p_m^{\mathrm{src}}=p_m^{\mathrm{tgt}}$, while for temporal transitions the
source and target marginals are computed from adjacent visits. We further
compute a $\chi^2$ test of dependence from a $2\times2$ contingency table.
To avoid data leakage, all EHR-derived co-occurrence and temporal-transition
statistics used to construct the global TKG are computed only from the training
split within each fold.

\paragraph{Evidence filtering.}
For each pair, we consolidate the co-occurrence and temporal statistics into
a compact evidence summary:
\begin{equation}
    \mathbf{a}_{ij}
    =
    [
    \mathbf{a}^{\mathrm{co}}_{ij};
    \mathbf{a}^{\mathrm{temp}}_{ij}
    ],
\end{equation}
where each channel-specific vector contains support, conditional probability,
PMI, and the $\chi^2$ significance score. Candidate pairs with insufficient
support, weak association, or non-significant dependence are removed before
LLM-based relation induction. This filtering step prevents assigning semantic
relations to arbitrary code pairs and ensures that the candidate set is
grounded in empirical EHR evidence.

\paragraph{Type-constrained LLM relation induction.}
Each remaining candidate pair is passed to a type-constrained LLM relation
induction module. For a pair $(c_i,c_j)$, the allowed relation set is selected
according to the source and target code types:
\begin{equation}
    \mathcal{R}_{ij}
    =
    \mathcal{R}_{\phi(c_i),\phi(c_j)} .
\end{equation}
For example, diagnosis--medication pairs may use relations such as
\texttt{treats}, \texttt{prevents}, \texttt{contraindicated\_for}, and
\texttt{causes\_adverse\_event}, while diagnosis--diagnosis pairs may use
relations such as \texttt{causes}, \texttt{risk\_factor\_for},
\texttt{leads\_to}, and \texttt{complicates}. Ambiguous cases are handled by
conservative abstention labels such as \texttt{no\_significant\_relation} and
\texttt{cannot\_decide}.

The LLM prompt includes the two code identifiers, code names, clinical types,
parent categories, marginal frequencies, the EHR-derived evidence summary
$\mathbf{a}_{ij}$, a short glossary of the statistical metrics, and the
allowed type-specific relation set $\mathcal{R}_{ij}$. The LLM is instructed
to return a structured output containing a directed relation label, an
oriented triple, a confidence score, and a short clinical rationale. We use
the confidence score and format checks for quality control, while the final
graph stores the selected relation type through $\psi(e)$ and the textual
rationale as the edge text $\eta_e$.

\paragraph{Node and edge textual attributes.}
After relation induction, we enrich the graph with textual attributes. For
each node $v\in\mathcal{V}$, we attach a concise clinical description $d_v$
generated using a type-specific prompt. The description summarizes the
medical meaning of the code, such as disease presentation for diagnoses,
therapeutic role for medications, or clinical purpose for procedures.

For each retained edge $e=(u,v)\in\mathcal{E}$, we attach the LLM-generated
relation rationale $\eta_e$ as the edge text. Thus, the text-attributed graph
contains both node-level concept semantics and edge-level relational
semantics:
\begin{equation}
    v \mapsto d_v,
    \qquad
    e \mapsto \eta_e .
\end{equation}
The statistical evidence is used during graph construction and relation
induction, but in the current formulation the edge attribute used by the
downstream model is the textual rationale $\eta_e$.

\paragraph{Directionality.}
The induced edges are directed and preserve their clinical orientation. For
example, if the LLM assigns a relation indicating that a medication treats a
diagnosis, the resulting edge direction follows the induced semantic triple.
For message passing, we add inverse computational edges during local graph
materialization to support bidirectional information flow in the GNN. They do not change the semantic
direction of the original global KG relation.

\paragraph{Use in patient-specific temporal graph construction.}
The global graph $\mathcal{G}$ serves as the source graph from which
patient-specific temporal graphs are extracted. Given a patient $p$ and a
prediction cutoff time $t^\ast$, only codes observed up to $t^\ast$ are used
as root nodes. The selected neighborhoods of these roots provide relational KG
context for the patient graph. The node descriptions $\{d_v\}$ and edge
rationales $\{\eta_e\}$ provide the textual semantics used to initialize or
generate the graph-contextualized code and KG soft prompts in \model{}.

\subsection{LLM-Based Node Description Construction}
\label{app:node_description_construction}

In addition to relation-typed edges, we enrich each medical concept node with
a textual description following the node-level enrichment strategy of
\citet{kerdabadi2026text}. For each node $v\in\mathcal{V}$, corresponding to a
medical code $c$, we generate a concise clinical description $d_v$ using a
type-specific LLM prompt:
\begin{equation}
    d_v
    =
    \operatorname{LLM}
    \left(
    \mathcal{P}_{\mathrm{desc}}
    (c,\phi(v))
    \right),
\end{equation}
where $\phi(v)\in\mathcal{T}_c$ denotes the code type, and
$\mathcal{P}_{\mathrm{desc}}$ is a prompt template specialized for diagnosis,
medication, and procedure codes.

The prompt provides the code identifier, code name, clinical type, and
available parent category information, and instructs the LLM to generate a
single clinically focused paragraph. The type-specific guidance asks for
different information depending on the code category: diagnosis descriptions
summarize the condition definition, typical presentation, risk factors, and
clinical management; procedure descriptions summarize clinical purpose,
indications, contraindications, and peri-procedural considerations; and
medication descriptions summarize pharmacologic class, mechanism, indications,
adverse effects, and common interaction patterns. To avoid unsupported or
overly specific content, the prompt asks the LLM to avoid patient-level facts,
URLs, dosage thresholds, hospital policies, and uncertain claims.

The LLM output is parsed as structured JSON and the generated description is
attached to the corresponding node as its textual attribute:
\begin{equation}
    v \mapsto d_v .
\end{equation}
These node descriptions provide fixed text-derived clinical semantics used by
\model{} to initialize code representations and construct graph-aware soft
prompts. Together with the edge rationales $\eta_e$ generated during the relation
induction, they make $\mathcal{G}$ a relation-typed and text-attributed medical
KG.

\begin{figure*}[t]
    \setlength{\abovecaptionskip}{2pt} % Adjust the space above the caption
    \begin{center}
    %\centering
    \includegraphics[width=0.9\linewidth]{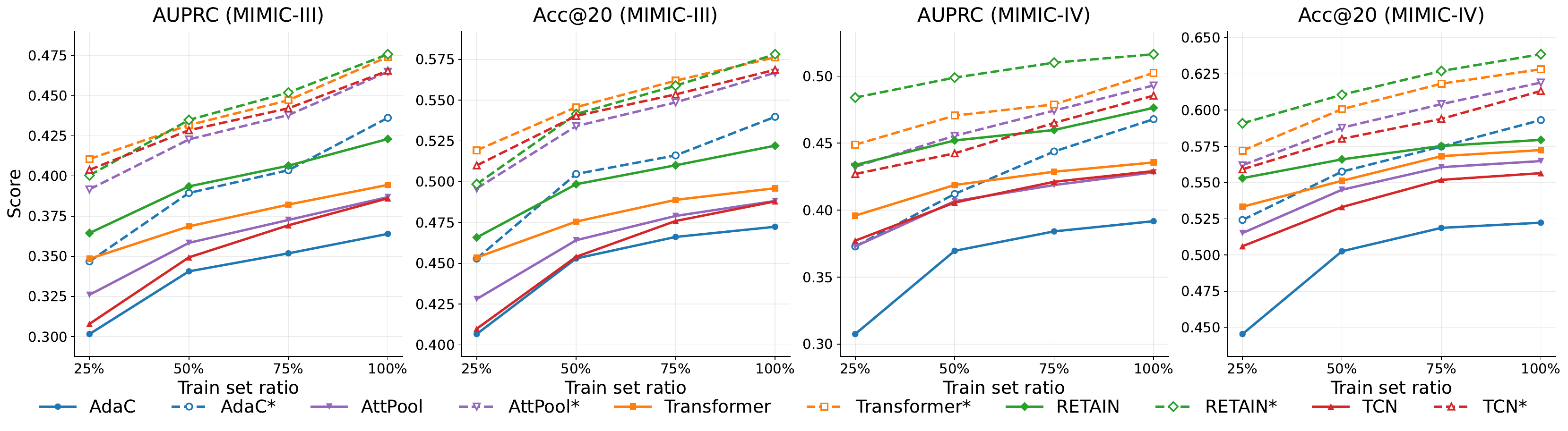}
    \caption{
    Data insufficiency analysis on MIMIC-III and MIMIC-IV under different training-set ratios. 
    We compare each backbone model with its \model{}-enhanced variant across AUPRC and Acc@20. 
    Solid lines denote the original backbone models, while dashed lines with hollow markers denote the corresponding models augmented with \model{}. Across datasets, metrics, and backbone architectures, \model{} consistently improves performance, demonstrating its robustness under limited supervision.
    }
    \label{fig:data_insufficiency}
    \end{center}
    \vspace{-5mm}
\end{figure*}

\subsection{Implementation Details}
\label{app:implementation_details}

We map ICD diagnosis/procedure codes to CCS categories and medications from National Drug Codes (NDC) to the Anatomical Therapeutic Chemical (ATC) classification. We use \texttt{Llama-3.2-1B}~\citep{grattafiori2024llama3,meta2024llama32_1b} as the frozen LLM backbone. All reported results are averaged over five-fold experiments using the same preprocessing and evaluation protocol.

\paragraph{Hyperparameters.}
The heterogeneous GNN uses GAT-style message passing with hidden dimension $128$, $4$ layers, $2$ attention heads, dropout $0.4$, pooled-code patient representation, and bidirectional KG message passing. We run the frozen LLM in bfloat16 precision and do not update its tokenizer, embedding table, or backbone parameters. The trainable components include the heterogeneous GNN, residual code and relation embeddings, text-semantic projection layers, KG-context pooling module, soft-token projections, GNN--LLM fusion layer, downstream EHR backbone, and prediction head. We use graph-aware mixed prompts with hard text tokens for the task and visit-code scaffold, and two soft tokens per observed code: one code soft token and one KG-context soft token.

We train with Adam using a learning rate of $5\times10^{-4}$, weight decay $0$, and batch size $256$. In the first-stage graph-budgeting phase, we warm up the proxy GNN for $120$ epochs and then train the sequential RL selector for up to $50$ epochs with early stopping after at least $5$ epochs, patience $5$, and minimum improvement $10^{-5}$ on the mean RL reward. The RL selector uses additive transitions with minimum and maximum decision steps of $2$ and $5$, respectively. The KG cost weights are $\omega_C=0.05$ for selected nodes and $\omega_E=0.5$ for selected edges, with cost penalty $\lambda_{\mathrm{cost}}=10^{-4}$. The actor, critic, and entropy weights are $0.1$, $0.1$, and $0.02$, respectively. During RL training, actions are sampled for exploration, while validation and inference use greedy selection. The task model is frozen during RL selector training, and selected patient graphs are exported for second-stage full \model{} training.

\paragraph{Two-stage training pipeline.}
Algorithm~\ref{alg:refine_training} summarizes the two-stage training pipeline of \model{}. We train \model{} in two stages to separate discrete graph-budget learning from the more expensive LLM-based semantic refinement. In the first stage, we train a personalized graph-budgeting policy using a GNN-only proxy predictor, avoiding repeated LLM rollouts during discrete graph selection. For each patient $p$, let $\mathcal{S}_p=\bigcup_{t\leq t^\ast}V_{p,t}$ denote the set of distinct observed code roots up to the prediction cutoff. The policy selects from a discrete action grid $a=(K,\mathbf{m})$, where $K$ is the KG expansion depth and $\mathbf{m}=(m_1,\ldots,m_K)$ specifies the per-hop fanout. The grid includes the no-expansion action $a_0=(0,\emptyset)$ and bounded expansions up to $K_{\max}=2$.

In Algorithm~\ref{alg:refine_training}, $A_p$ denotes the current action map that assigns each root code $c\in\mathcal{S}_p$ to an expansion action $a\in\mathcal{A}$. At RL step $\ell$, $\mathcal{U}_\ell$ denotes the feasible macro-action set, where each macro-action $b=(c,a)$ updates the expansion action for one distinct observed code root. The operator $\textsc{Expand}(\mathcal{S}_p,A_p,\mathcal{G})$ materializes the selected patient graph $\mathcal{G}^{\mathrm{sel}}_p$ from the global TKG $\mathcal{G}$ according to the action map $A_p$. We use $C_{\mathrm{kg}}(A_p)$ to denote the induced KG cost, $B_{\mathrm{kg}}$ for the graph budget, $L_{\max}$ for the maximum number of RL decisions, and $\mathcal{M}$ for the exported selected-graph cache. The process terminates when the policy selects the stop action, reaches $L_{\max}$, or exhausts the graph budget.

The reward is computed from the downstream GNN-only prediction loss improvement relative to the no-expansion graph, with an additional penalty for selected KG nodes and edges. During training, actions are sampled from the actor for exploration; during validation and inference, we use greedy selection. The best selected patient-specific graphs are exported and cached in $\mathcal{M}$.

In the second stage, we fix the selected patient graphs and train the full \model{} architecture for next-visit diagnosis prediction. For each patient $p$, $\textsc{GNNEnc}(\cdot)$ produces structural representations $H_p^G$, $\textsc{LLMRefine}(\cdot)$ produces frozen-LLM-refined semantic representations $H_p^L$, and $\textsc{Fuse}(\cdot)$ combines them into personalized code representations $Z_p=\{z_{p,t,c}\}$. The LLM remains frozen, and only the trainable \model{} components are updated, including the heterogeneous GNN, residual code and relation embeddings, text-semantic projection layers, KG-context pooling module, soft-token projections, GNN--LLM fusion layer, downstream EHR backbone, and prediction head. This design separates graph-budget learning from LLM-based semantic refinement while preserving patient-personalized code representations from the selected graph context.

\subsection{KG-Context Pooling and Prompt Design}
\label{app:kgpool_prompt_design}

For each patient--code pair $(p,c)$, REFINE constructs a selected
code-specific KG subgraph $\mathcal{G}_{p,c}^{\mathrm{kg}}$. The goal of
$\operatorname{KGPool}(\cdot)$ is to summarize this selected relational context
as a single KG-context soft token $\mathbf{k}_{p,c}$ that can be inserted into
the frozen LLM prompt.

We considered several implementations, including mean pooling, gated attention,
and relation-aware aggregation over selected neighbor nodes and KG edges. The
best-performing implementation uses a lightweight GNN-style aggregator. For
each selected KG edge $(a,b)\in\mathcal{E}_{p,c}^{\mathrm{kg}}$, we first
combine the LLM-branch representation of the neighbor node with the
corresponding relation/rationale-aware edge representation. These edge-aware
messages are then aggregated over the selected code-specific subgraph to
produce a compact context vector, which is projected into the LLM hidden space
as $\mathbf{k}_{p,c}$. This design preserves local relational information
while avoiding the need to insert every selected KG triple as a separate prompt
object.

We also explored a more explicit relational prompt format inspired by recent
relational LLM prompting work, where relational structures are serialized in a
JSON-like format and graph-derived embeddings are inserted as soft prompt
tokens. Rel-LLM, for example, argues that flat text serialization can obscure
relational structure, introduce redundancy, and exceed LLM context limits, and
instead uses graph encoders and structured prompts to preserve relational
information. Following this intuition, we adapted a triplet-style prompt in which selected
KG evidence was represented as soft-tokenized triples of the form
$(\mathrm{node},\mathrm{edge},\mathrm{node})$ inside our JSON visit-code
hierarchy. However, this design substantially increased prompt length and the
number of soft tokens, especially for patients with many observed codes and
multi-hop KG expansions. Empirically, this led to higher memory and computational costs and weaker performance than the compact KGPool design. Therefore, REFINE
uses KGPool to summarize each selected code-specific KG neighborhood into one
KG-context token, balancing relational expressiveness and prompt efficiency.

\subsection{Value-of-Information KG Selection}
\label{app:voi_selection}

As an alternative to fixed KG expansion and sequential RL budgeting, we consider a learned value-of-information (VoI) selection strategy. The goal is to predict the KG expansion action expected to provide the largest predictive benefit relative to a no-expansion baseline. We use the same discrete hop--fanout action grid as the RL policy: $(0,\emptyset)$, $(1,(1))$, $(1,(2))$, $(1,(5))$, $(1,(10))$, $(2,(1,1))$, $(2,(2,1))$, $(2,(2,2))$, $(2,(5,1))$, and $(2,(5,5))$.

VoI uses a learned critic to score candidate expansion actions at the visit level. During training, sparse utility targets are constructed from training labels. Let $\mathcal{G}_{p}^{0}$ denote the baseline patient graph in which all visits use the no-expansion action. For a candidate action $a\in\mathcal{A}$ at visit $t$, let $\mathcal{G}_{p}^{(t,a)}$ denote the graph in which visit $t$ uses action $a$ while all other visits remain at no expansion. Given a prediction loss $\mathcal{L}_{p}(\cdot)$, the target utility is defined as
\begin{equation}
    \Delta_{p,t}(a)
    =
    \mathcal{L}_{p}(\mathcal{G}_{p}^{0})
    -
    \mathcal{L}_{p}(\mathcal{G}_{p}^{(t,a)}).
\end{equation}

These sparse utility targets supervise the VoI critic, which predicts a utility score $\widehat{\Delta}_{p,t}(a)$ for each candidate action. At validation and test time, ground-truth labels are not used for KG selection. Instead, the selected VoI action is
\begin{equation}
    \hat{a}_{p,t}
    =
    \arg\max_{a\in\mathcal{A}}
    \widehat{\Delta}_{p,t}(a).
\end{equation}

Unlike fixed expansion, VoI adapts KG expansion across visit contexts and performs label-free selection at inference. However, its visit-level decisions are made independently rather than sequentially conditioning on KG context already selected for other code roots. In contrast, the RL policy makes sequential code-level decisions conditioned on the patient state, remaining budget, and previously selected KG context, enabling finer-grained personalized graph construction.

\subsection{RQ5: Data Insufficiency Analysis}
\label{app:appendix_data_insufficiency}

To evaluate the robustness of \model{} under limited supervision, we conduct a data insufficiency analysis by varying the fraction of the available training set. This setting simulates practical clinical scenarios where labeled EHR data are limited. As shown in Figure~\ref{fig:data_insufficiency}, \model{} consistently improves downstream prediction performance across different training-set ratios on both MIMIC-III and MIMIC-IV. The improvements are observed across multiple backbone models, including AdaCare, AttPool, Transformer, RETAIN, and TCN, indicating that \model{} functions as a general plug-in refinement module rather than being specific to one architecture. The gains are especially meaningful in low-data regimes, where base EHR models have limited supervision to learn reliable representations for sparse and rare medical codes. By incorporating patient-specific graph context and semantic refinement, \model{} provides additional inductive bias that helps improve both AUPRC and Acc@20. Importantly, the enhanced models remain stronger than their base counterparts even at the full-data setting, suggesting that \model{} provides complementary information beyond what is learned from labeled sequences alone. Overall, these results show that \model{} is robust to data insufficiency and can improve EHR prediction performance across datasets, architectures, and supervision levels.

\begin{figure}[t]
    \centering
    \includegraphics[width=\linewidth]{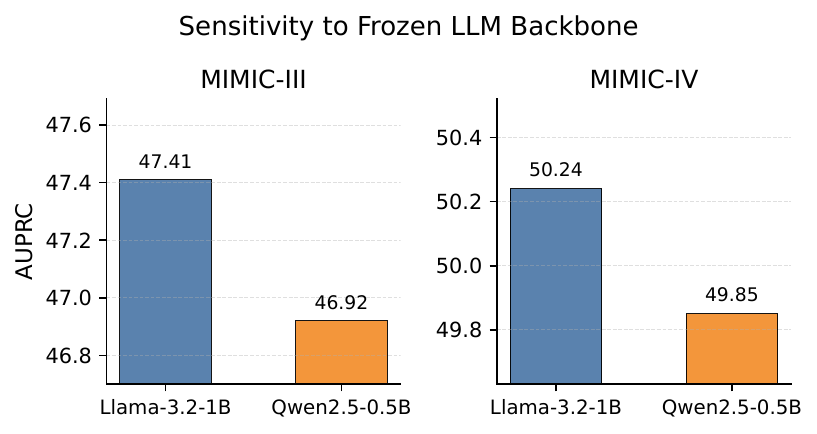}
    \caption{
    Sensitivity of \model{} to the frozen LLM backbone on MIMIC-III and MIMIC-IV. 
    \model{} remains stable when replacing \texttt{Llama-3.2-1B} with the smaller \texttt{Qwen2.5-0.5B}.
    }
    \label{fig:llm_backbone_sensitivity}
\end{figure}

\subsection{Sensitivity to LLM Backbone}
\label{app:llm_backbone_sensitivity}

We evaluate the sensitivity of \model{} to the frozen LLM backbone by comparing \texttt{Llama-3.2-1B}~\citep{grattafiori2024llama3,meta2024llama32_1b} with the smaller \texttt{Qwen2.5-0.5B}~\citep{hui2024qwen2,qwen2024qwen25_05b}. As shown in Figure~\ref{fig:llm_backbone_sensitivity}, replacing \texttt{Llama-3.2-1B} with \texttt{Qwen2.5-0.5B} leads to a moderate decrease in AUPRC on both MIMIC-III and MIMIC-IV, but the overall performance remains competitive. This suggests that \model{} can benefit from its graph-contextual refinement mechanism across different small frozen LLM backbones. However, this analysis is limited to relatively small models, and future work should evaluate larger LLM backbones to better understand how backbone capacity affects semantic refinement and downstream EHR prediction.

\end{document}